\documentclass{article}

\usepackage[utf8]{inputenc}
\usepackage[T1]{fontenc}
\usepackage{times}
\usepackage{helvet}
\usepackage{courier}

\usepackage{amsmath}
\usepackage{amssymb}
\usepackage{amsfonts}

\usepackage{array}
\usepackage{booktabs}
\usepackage{makecell}
\usepackage{multirow}
\usepackage{tabularx}

\usepackage{graphicx}
\usepackage{xcolor}
\usepackage{caption}

\usepackage{algorithm}
\usepackage{algorithmic}

\usepackage{enumitem}
\usepackage{float}
\usepackage{microtype}

\usepackage{url}
\usepackage[colorlinks=true, linkcolor=red, citecolor=green, urlcolor=cyan]{hyperref}

\usepackage[numbers,square,sort&compress]{natbib}

\usepackage{arxiv}

\newcommand{\model}{DynImmune-BERT}
\newcommand{\R}{\mathbb{R}}
\definecolor{refred}{RGB}{230,0,0}
\newcommand{\redref}[1]{\textcolor{refred}{\ref{#1}}}

\title{\model: Dynamic Immune Repertoire Modeling with Neural ODE Driven Continuous Transformers}

\author{
    Rong Fu \\
    Independent Researcher \\
    Corresponding author \and
    Yongtai Liu \\
    Independent Researcher \and
    Xiaowen Ma \\
    Independent Researcher \and
    Haoyu Zhao \\
    Independent Researcher \and
    Shuo Yin \\
    Independent Researcher \and
    Yiqing Lyu \\
    Independent Researcher \and
    Long Zhang \\
    Independent Researcher \and
    Wangyu Wu \\
    Independent Researcher
}

\renewcommand{\headeright}{}
\renewcommand{\undertitle}{}
\renewcommand{\shorttitle}{\model}

\hypersetup{
    pdftitle={DynImmune-BERT: Dynamic Immune Repertoire Modeling with Neural ODE Driven Continuous Transformers},
    pdfsubject={Quantitative Biology - Quantitative Methods (q-bio.QM); Machine Learning (cs.LG)},
    pdfauthor={Rong Fu, Yongtai Liu, Xiaowen Ma, Haoyu Zhao, Shuo Yin, Yiqing Lyu, Long Zhang, Wangyu Wu},
    pdfkeywords={immune repertoire, T cell receptor, Neural ODE, continuous Transformer, cancer detection},
    pdfstartview={FitH}
}

\begin{document}

\maketitle

\begin{abstract}
Longitudinal T cell receptor repertoires contain signals of clonal expansion, contraction, disappearance, and reappearance after immune perturbation. Static repertoire language models usually summarize a sample as a bag of sequences, so the sampling interval, sequencing depth, and clone presence pattern are only weakly represented. This paper presents \model, a continuous time repertoire model for patient level immune status prediction. The method combines depth adaptive centered log ratio initialization, clone presence gated Neural ordinary differential equation dynamics, bounded neighborhood self attention, event based state restart, and a hybrid transport objective that supervises dominant and rare clone mass. A low rank meta adapter initializes reappearing clonotypes while keeping the parameter count independent of the number of observed clones. The evaluation separates literature reported baselines from internally controlled temporal comparisons, reports uncertainty for small external cohorts, adds calibration and threshold diagnostics, and visualizes latent clone trajectories and attention neighborhoods. The results indicate that event aware temporal modeling can complement strong static encoders when longitudinal repertoire structure is available, while small external cohorts and protocol differences require cautious interpretation.
\end{abstract}


\section{Introduction}
Adaptive immune repertoires preserve a temporal record of antigen exposure, clonal selection, immune memory, and disease related remodeling. In T cell receptor (TCR) sequencing, clinically relevant information may appear not only in the amino acid sequence of a clonotype, but also in whether that clonotype expands, contracts, disappears, persists, or reappears across irregular sampling times. Longitudinal studies have connected repertoire dynamics with infection history, aging, inflammatory disease activity, therapy response, and cancer associated immune remodeling \cite{niu2020longitudinal,sun2022longitudinal,minervina2021longitudinal,mitchell2022temporal,chen2022longitudinal,servaas2021longitudinal,kjaer2025low,briggs2025longitudinal}. These observations motivate computational models that represent an immune repertoire as an evolving patient trajectory.

Deep learning has substantially improved sequence level and repertoire level analysis. Motif based predictors, multiple instance learning, contrastive sequence encoders, graph models, and pretrained protein language models have strengthened antigen prediction, immune status prediction, and cancer related repertoire classification \cite{beshnova2020novo,xu2022deeplion,qian2024deeplion2,sidhom2021deeptcr,zhang2024berttcr,ni2025methpriorgcn,dounas2024learning,im2025sequence,fang2024large,leem2022deciphering}. Earlier repertoire classifiers also showed that handcrafted features and sequence aware neural encoders can be informative when cohorts are carefully defined \cite{lu2021deep,cai2024deep,mosch2019machine}. Recent cancer detection studies further suggest that circulating TCR signals can support noninvasive immune monitoring \cite{li2025circulating,zaslavsky2025disease,o2024reading}. However, most existing models aggregate clonotypes within one sample or compare samples independently. This design is effective for static classification, but it leaves a mismatch between the biological process, which is sparse, heavy tailed, and event driven, and the computational representation, which is usually time independent.

Longitudinal TCR repertoires are usually sampled at irregular intervals that may span weeks or months. We use Neural ODEs as a mathematical framework for propagating latent states between unequally spaced observations without imposing a fixed discretization grid \cite{chen2018neural,vaswani2017attention,zhang2021continuous,chen2023contiformer,zhang2025diffode}. This use of ODE integration should be understood as latent interpolation over irregular time stamps, not as a mechanistic claim that clonal dynamics are biologically smooth. Direct application to TCR data is still nontrivial. Repertoire measurements are compositional because clone frequencies are constrained by sequencing depth. Many clones are absent at most time points, and absence may reflect true biological disappearance, limited sampling depth, or technical censoring. Dominant clones can mask rare but informative clonotypes. Clone birth and death also violate the assumption that all latent states move smoothly under one continuous vector field.

This work studies the problem of learning a temporally aware repertoire representation for cancer related immune status prediction. The key question is how to encode a subject's irregular TCR trajectory while preserving clone presence events, compositional stability, and supervision from both high abundance and low abundance clones. We address this question with \model, a Neural ODE driven continuous Transformer that combines event aware clone dynamics with bounded neighborhood attention and transport based distribution matching.

Our contributions are as follows. First, we formulate immune repertoire classification as a patient level trajectory problem in which observation time, clone read count, sequencing depth, and clone presence are explicit variables. Second, we introduce an event aware continuous Transformer that uses stabilized compositional states, bounded clone neighborhoods, implicit integration, and low rank restart for reappearing clonotypes. Finally, we combine internally controlled temporal comparisons with uncertainty, calibration, and trajectory diagnostics so that dynamic behavior can be examined alongside aggregate classification metrics.

\section{Related Work}
\subsection{Immune Repertoire Learning}
TCR repertoire sequencing has been used to study infection, autoimmunity, cancer immunotherapy, and early cancer detection \cite{chiffelle2020t,pai2021high,li2025circulating}. Early machine learning systems often depended on motifs, summary statistics, or clone sharing. DeepCAT, DeepLION, DeepLION2, DeepTCR, MethPriorGCN, and BertTCR showed that neural encoders can extract stronger repertoire level signals from CDR3 sequences and clone bags \cite{beshnova2020novo,xu2022deeplion,qian2024deeplion2,sidhom2021deeptcr,ni2025methpriorgcn,zhang2024berttcr}. Multiple instance networks and attention based encoders also provide useful baselines for sample level classification \cite{kim2022multiple,shao2021transmil,zhu2023biformer}. These models establish strong static references, but most of them do not represent irregular observation intervals as part of the learning problem.

\subsection{Continuous Time Sequence Models}
Neural ODEs define hidden state evolution with a learned vector field and support adaptive integration between observations \cite{chen2018neural}. Continuous attention and continuous Transformer models extend this idea to sequence modeling with irregular time stamps \cite{zhang2021continuous,chen2023contiformer,tong2025neural}. Biomedical trajectories often contain abrupt clinical or biological events, so event localization and state restart are important for avoiding unrealistic smooth interpolation. \model\ adapts this principle to repertoire modeling by treating clone disappearance and reappearance as observed events that interrupt continuous evolution.

\subsection{Transport Objectives and Efficient Adaptation}
Optimal transport losses compare probability distributions while preserving geometric structure \cite{mensch2020online,mialon2020trainable}. Entropic transport is effective on restricted supports, whereas sliced Wasserstein distances scale to diffuse distributions \cite{nguyen2023energy,heitz2021sliced}. Low rank adapters provide parameter efficient adaptation and can be controlled by meta networks \cite{zhang2023lora,zhang2024autolora,hayou2024lora+}. \model\ combines these ideas to supervise dominant clone mass and rare clone mass while restarting reappearing clones through a compact adapter.

\begin{figure*}[!t]
\centering
\includegraphics[width=0.92\textwidth]{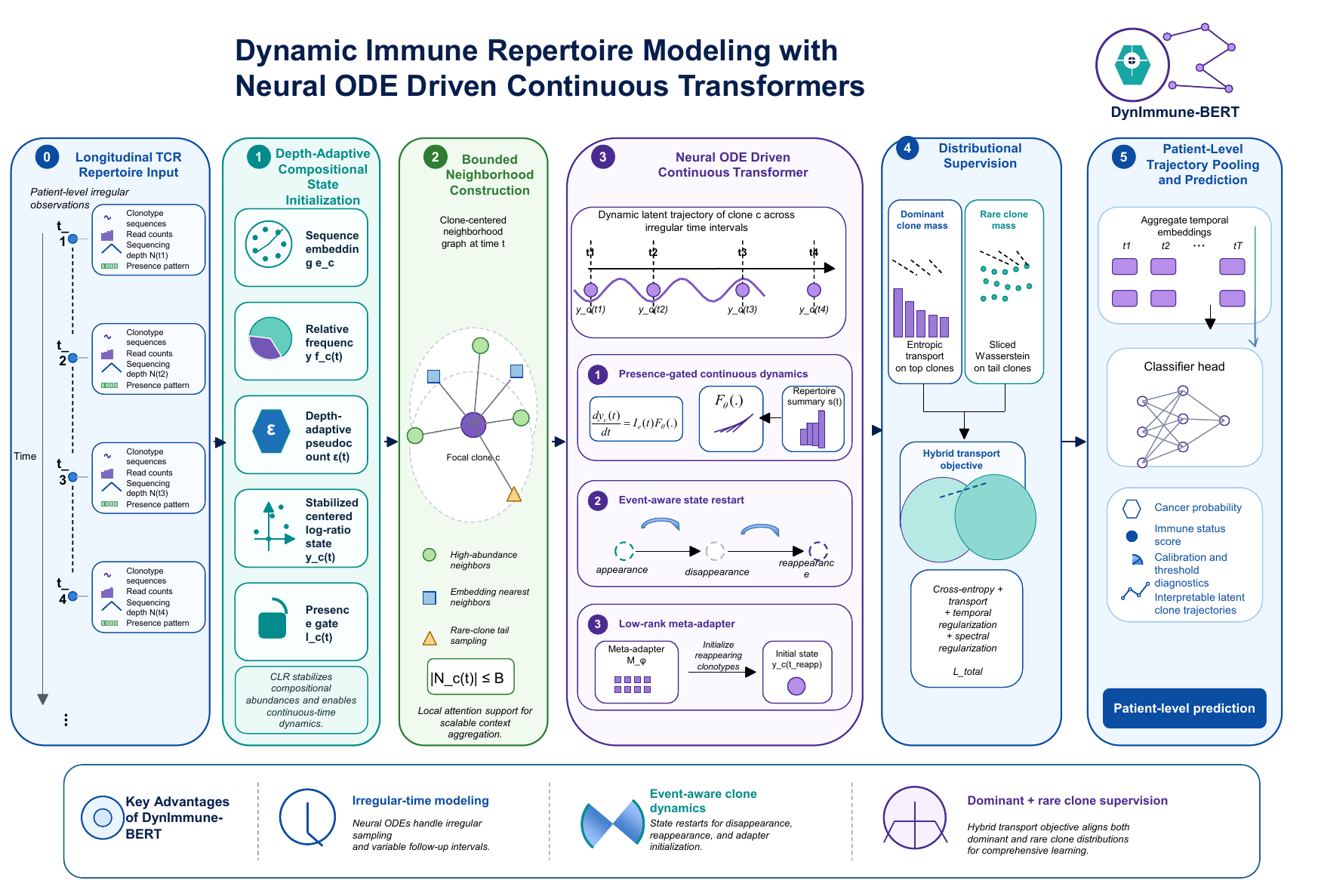}
\caption{Overview of \model. Input TCR sequences, read counts, sampling times, and patient labels are processed through stabilized compositional state initialization, bounded neighborhood attention, event aware restart, hybrid transport supervision, and patient level trajectory pooling to produce cancer probability and calibrated health scores.}
\label{fig:framework}
\end{figure*}

\section{Methodology}
\subsection{Formal Problem Formulation}
For subject \(p\), the observed longitudinal repertoire is defined as
\begin{equation}
\mathcal{D}_{p}=\{(t_{p,k},\mathcal{C}_{p,k},\mathbf{n}_{p,k},y_p)\}_{k=1}^{K_p}.
\label{eq:subject_data}
\end{equation}
where \(t_{p,k}\) is the \(k\)-th sampling time, \(\mathcal{C}_{p,k}\) is the observed clonotype set, \(\mathbf{n}_{p,k}\) contains clone read counts, \(y_p\in\{0,1\}\) is the patient label, and \(K_p\) is the number of observed time points.

The model estimates a patient level prediction by
\begin{equation}
\widehat{y}_p=g_{\psi}\left(\{\mathbf{z}_{p}(t_{p,k})\}_{k=1}^{K_p}\right).
\label{eq:prediction_problem}
\end{equation}
where \(g_{\psi}\) is the patient classifier and \(\mathbf{z}_{p}(t_{p,k})\) is a time dependent repertoire representation produced by the dynamic encoder.

All evaluation splits satisfy
\begin{equation}
\mathcal{P}_{\mathrm{train}}\cap\mathcal{P}_{\mathrm{val}}\cap\mathcal{P}_{\mathrm{test}}=\varnothing.
\label{eq:patient_split}
\end{equation}
where \(\mathcal{P}_{\mathrm{train}}\), \(\mathcal{P}_{\mathrm{val}}\), and \(\mathcal{P}_{\mathrm{test}}\) are patient sets, so all samples from a patient remain in one split.

\subsection{Depth Adaptive Compositional State}
Let \(n_c(t)\) be the read count of clone \(c\) at time \(t\). The total sequencing depth is
\begin{equation}
N(t)=\sum_{m\in\mathcal{C}_t}n_m(t).
\label{eq:depth}
\end{equation}
where \(N(t)\) is the number of reads at time \(t\), \(\mathcal{C}_t\) is the observed support at time \(t\), and \(n_m(t)\) is the read count of clone \(m\).

The relative clone abundance is
\begin{equation}
f_c(t)=\frac{n_c(t)}{N(t)}.
\label{eq:relative_frequency}
\end{equation}
where \(f_c(t)\) is the normalized frequency of clone \(c\), and the denominator is the sequencing depth defined in \eqref{eq:depth}.

To reduce instability from zero counts, we use a depth adaptive pseudocount
\begin{equation}
\varepsilon(t)=\frac{\kappa}{N(t)+\kappa}.
\label{eq:pseudocount}
\end{equation}
where \(\varepsilon(t)\) is the smoothing term and \(\kappa>0\) controls how strongly low depth observations are regularized.

The initial clone state is a stabilized centered log ratio transform:
\begin{equation}
y_c(t)=\ln(f_c(t)+\varepsilon(t))-\frac{1}{|\mathcal{S}_t|}\sum_{m\in\mathcal{S}_t}\ln(f_m(t)+\varepsilon(t)).
\label{eq:clr_state}
\end{equation}
where \(y_c(t)\) is the compositional log state, \(\mathcal{S}_t\) is the union of observed and retained candidate clones at time \(t\), and \(|\mathcal{S}_t|\) is the support size used for centering.

Clone presence is represented by
\begin{equation}
I_c(t)=\mathbb{1}[n_c(t)>0].
\label{eq:presence}
\end{equation}
where \(I_c(t)\) equals one when clone \(c\) is observed and equals zero when the clone is absent at time \(t\).

\subsection{Bounded Neighborhood Construction}
Each clone is associated with a sequence embedding \(e_c\in\R^d\). The dynamic attention neighborhood is built as
\begin{equation}
\mathcal{N}_c(t)=\mathcal{N}^{\mathrm{freq}}_c(t)\cup\mathcal{N}^{\mathrm{emb}}_c(t)\cup\mathcal{N}^{\mathrm{tail}}_c(t).
\label{eq:neighbor_union}
\end{equation}
where \(\mathcal{N}^{\mathrm{freq}}_c(t)\) contains high abundance clones, \(\mathcal{N}^{\mathrm{emb}}_c(t)\) contains nearest sequence embedding neighbors, and \(\mathcal{N}^{\mathrm{tail}}_c(t)\) contains sampled low abundance candidates.

The bounded size condition is
\begin{equation}
|\mathcal{N}_c(t)|\leq B.
\label{eq:bounded_neighbor}
\end{equation}
where \(B\) is the maximum number of neighbors used by each clone at one observation time.

The neighborhood bound is used primarily to control computation on repertoires with thousands of candidate clones. It reduces attention from full pairwise comparison to local aggregation while preserving three sources of context: abundance, sequence similarity, and low abundance exploration. In the ablation study, removing the neighborhood module causes only a modest AUC decline, but the bounded form keeps memory and inference latency practical for large clone supports.

Attention over the bounded neighborhood is computed as
\begin{equation}
a_{cj}^{(h)}(t)=
\frac{\exp(\ell_{cj}^{(h)}(t)-m_c^{(h)}(t))}
{\sum_{r\in\mathcal{N}_c(t)}\exp(\ell_{cr}^{(h)}(t)-m_c^{(h)}(t))}.
\label{eq:attention}
\end{equation}
where \(a_{cj}^{(h)}(t)\) is the head \(h\) attention weight from clone \(c\) to neighbor \(j\), \(\ell_{cj}^{(h)}(t)\) is the scaled query key logit, and \(m_c^{(h)}(t)=\max_{r\in\mathcal{N}_c(t)}\ell_{cr}^{(h)}(t)\) is used for numerical stability.

\subsection{Event Aware Continuous Dynamics}
Between two observed events, the clone state evolves according to
\begin{equation}
\frac{d y_c(t)}{dt}=I_c(t)\mathcal{F}_{\theta}\left(y_c(t),e_c,\mathcal{N}_c(t),\mathbf{s}(t),t\right).
\label{eq:event_ode}
\end{equation}
where \(\mathcal{F}_{\theta}\) is the continuous Transformer vector field, \(\mathbf{s}(t)\) is a pooled repertoire summary, and the presence gate \(I_c(t)\) prevents absent clones from drifting under the vector field.

The continuous flow is interrupted when a clone changes from absent to present. Such transitions can reflect true biological reappearance, increased sampling depth, or technical detection of a low frequency clone. For a restart event at \(t_i\), the uncentered restart proposal is
\begin{equation}
r_c(t_i)=\ln\varepsilon(t_i)+\Delta_{\phi}(e_c,t_i).
\label{eq:restart_proposal}
\end{equation}
where \(r_c(t_i)\) is the proposed restart state and \(\Delta_{\phi}\) is a meta low rank adapter controlled by clone embedding and time.

The support centered restart is
\begin{equation}
\mathbf{y}(t_i^+)=\Pi_{\mathcal{S}_{t_i}}\left(\mathbf{r}(t_i)\right).
\label{eq:centered_restart}
\end{equation}
where \(\mathbf{y}(t_i^+)\) is the post event state vector, \(\mathbf{r}(t_i)\) stacks restart proposals for active support \(\mathcal{S}_{t_i}\), and \(\Pi_{\mathcal{S}_{t_i}}\) subtracts the mean over \(\mathcal{S}_{t_i}\) so the centered log ratio constraint is restored after restart.

\subsection{Distributional Supervision}
The hidden clone state is mapped to a predicted clone distribution by
\begin{equation}
\widehat{p}_c(t)=\frac{\exp(\rho_c(t))}{\sum_{m\in\mathcal{S}_t}\exp(\rho_m(t))}.
\label{eq:pred_distribution}
\end{equation}
where \(\widehat{p}_c(t)\) is the predicted mass of clone \(c\), \(\rho_c(t)\) is a scalar readout from the dynamic state, and the denominator normalizes over the current support.

The hybrid transport loss is
\begin{equation}
\widetilde{W}(\mathbf{p},\widehat{\mathbf{p}})=
\omega_{\mathrm{top}}S_{\lambda}(\mathbf{p}_{\mathcal{T}},\widehat{\mathbf{p}}_{\mathcal{T}})
+\omega_{\mathrm{tail}}\mathrm{SW}_{L}(\mathbf{p}_{\mathcal{U}},\widehat{\mathbf{p}}_{\mathcal{U}}).
\label{eq:hybrid_transport}
\end{equation}
where \(S_{\lambda}\) is entropic transport on top clone set \(\mathcal{T}\), \(\mathrm{SW}_{L}\) is sliced Wasserstein distance with \(L\) random projections on tail sample \(\mathcal{U}\), and \(\omega_{\mathrm{top}}\) and \(\omega_{\mathrm{tail}}\) are nonnegative mixture weights.

The final training loss is
\begin{equation}
\mathcal{L}=\mathcal{L}_{\mathrm{CE}}+\lambda_W\widetilde{W}+\lambda_{\mathrm{temp}}\mathcal{R}_{\mathrm{temp}}+\lambda_{\mathrm{spec}}\mathcal{R}_{\mathrm{spec}}.
\label{eq:final_loss}
\end{equation}
where \(\mathcal{L}_{\mathrm{CE}}\) is cross entropy, \(\mathcal{R}_{\mathrm{temp}}\) penalizes unstable temporal codes, \(\mathcal{R}_{\mathrm{spec}}\) controls the vector field spectrum, and \(\lambda_W\), \(\lambda_{\mathrm{temp}}\), and \(\lambda_{\mathrm{spec}}\) are regularization weights.

\subsection{Stability and Complexity}
The event formulation is not meant to force every clone to follow a smooth biological path. Instead, it separates smooth latent evolution between observations from discrete state updates at detected clone presence changes. If the vector field is locally Lipschitz on a bounded state domain, then
\begin{equation}
\|\mathcal{F}_{\theta}(\mathbf{u},t)-\mathcal{F}_{\theta}(\mathbf{v},t)\|_2\leq L_F\|\mathbf{u}-\mathbf{v}\|_2.
\label{eq:lipschitz}
\end{equation}
where \(L_F\) is a finite Lipschitz constant, and \(\mathbf{u}\) and \(\mathbf{v}\) are two clone state vectors in the bounded integration domain.

The centered restart update is bounded by
\begin{equation}
\|\mathbf{y}(t_i^+)\|_2\leq \|\Pi_{\mathcal{S}_{t_i}}\|_2\left(\sqrt{|\mathcal{S}_{t_i}|}\,|\ln\varepsilon(t_i)|+\|\Delta_{\phi}(E,t_i)\|_2\right).
\label{eq:restart_bound}
\end{equation}
where \(E\) stacks clone embeddings in the active support, and the inequality shows that restart magnitude remains finite when sequencing depth and adapter output are bounded.

With bounded neighborhoods, the dominant forward cost satisfies
\begin{equation}
\mathcal{T}_{\mathrm{forward}}=\mathcal{O}(SCBd+SCd^2).
\label{eq:complexity}
\end{equation}
where \(S\) is the number of ODE solver steps, \(C\) is the number of candidate clones, \(B\) is the neighborhood bound, and \(d\) is the hidden dimension.

\begin{algorithm}[t]
\caption{\model\ training step}
\label{alg:dynimmune}
\begin{algorithmic}
\REQUIRE Longitudinal repertoire batch \(\{(t_k,\mathcal{C}_k,\mathbf{n}_k)\}_{k=1}^{K}\)
\REQUIRE Patient labels \(y\), dynamic encoder parameters \(\theta\), classifier parameters \(\psi\)
\ENSURE Prediction \(\widehat{y}\) and loss \(\mathcal{L}\)
\FOR{each observation time \(t_k\)}
\STATE Encode clone sequences as embeddings \(e_c\)
\STATE Compute \(f_c(t_k)\), \(\varepsilon(t_k)\), \(y_c(t_k)\), and \(I_c(t_k)\)
\STATE Build bounded neighborhoods from abundance, embedding similarity, and tail candidates
\ENDFOR
\FOR{each interval \([t_k,t_{k+1}]\)}
\STATE Integrate the gated vector field on smooth subintervals
\IF{a clone reappears}
\STATE Restart and recenter its state with \eqref{eq:centered_restart}
\ENDIF
\ENDFOR
\STATE Pool dynamic clone states into \(\mathbf{z}_p(t_K)\)
\STATE Predict \(\widehat{y}=g_{\psi}(\mathbf{z}_p(t_K))\)
\STATE Compute \(\mathcal{L}\) with \eqref{eq:final_loss}
\RETURN \(\widehat{y},\mathcal{L}\)
\end{algorithmic}
\end{algorithm}
\begin{figure*}[t]
\centering
\includegraphics[width=0.95\textwidth]{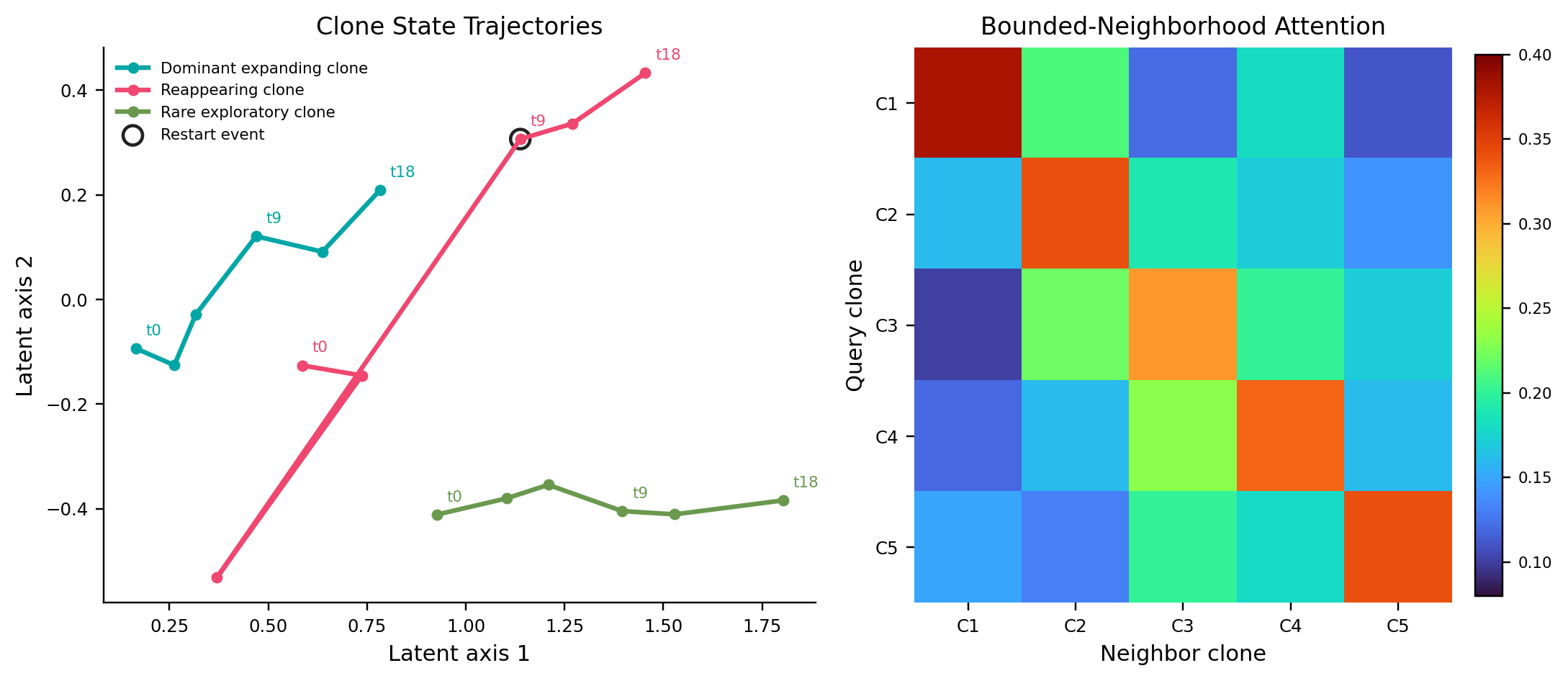}
\caption{Latent trajectory and attention diagnostics. The trajectory panel visualizes clone evolution and restart behavior, while the heatmap shows bounded neighborhood attention weights among representative clones.}
\label{fig:trajectory_attention}
\end{figure*}
\section{Experiments}
\subsection{Data and Comparison Protocol}
Experiments were conducted on a workstation running Ubuntu 20.04 with an AMD Ryzen R7 5700X CPU, 128GB RAM, and an NVIDIA RTX 4090 GPU (24GB VRAM). The software stack included Python 3.9.7, PyTorch 1.12.1+cu102, NumPy 1.7.1, pandas 1.3.4, and scikit-learn 0.24.2. Neural ODE modules were implemented using \textit{torchdiffeq} with implicit Runge-Kutta solvers for stable event handling. The evaluation separates three sources of evidence.  Table~\redref{tab:protocol} reflect different cohort definitions, preprocessing pipelines, and evaluation splits. Internally controlled temporal comparisons use the same patient level split, preprocessing, and evaluation script for \model\ variants and temporal controls. External cancer detection is reported as preliminary evidence because several disease subsets are small. Table~\redref{tab:protocol} summarizes the protocol.

\begin{table*}[t]
\centering
\caption{Evaluation protocol summary.}
\label{tab:protocol}
\resizebox{0.66\textwidth}{!}{%
\begin{tabular}{p{2.9cm}p{3.4cm}p{4.6cm}p{4.2cm}p{2.2cm}}
\toprule
Component & Source of numbers & Split and processing rule & Interpretation & Main output \\
\midrule
Literature context & Reported values from cited static TCR and MIL papers & Original protocols of cited reports are retained & Context for prior performance ranges & Table~\redref{tab:main_comparison} \\
Internal temporal controls & Same \model\ evaluation interface & Patient level split with all time points grouped before assignment & Matched comparison of temporal designs & Table~\redref{tab:temporal_controls} \\
External detection check & Public or cited repertoire cohorts summarized at disease level & Patient level separation and disease specific reporting & Small cohort generalization check & Table~\redref{tab:external_validation} \\
Threshold and calibration analysis & Validation scores and held out prediction files & Model selection and threshold selection are logged before test evaluation & Decision stability and probability reliability & Figs.~\redref{fig:youden} and~\redref{fig:calibration} \\
\bottomrule
\end{tabular}}
\end{table*}

The key hyperparameters were fixed before final test reporting. We used \(B=96\) bounded neighbors, \(\kappa=0.5\), \(\omega_{\mathrm{top}}=0.65\), \(\omega_{\mathrm{tail}}=0.35\), \(\lambda_W=0.2\), \(\lambda_{\mathrm{temp}}=10^{-3}\), and \(\lambda_{\mathrm{spec}}=10^{-4}\). The ODE solver used implicit Runge Kutta style steps with relative tolerance \(10^{-4}\) and absolute tolerance \(10^{-6}\). Hyperparameter values are reported alongside the notation to support independent implementation.

\subsection{Comparative Analysis with State-of-the-Art Methods}
Table~\redref{tab:main_comparison} reports lung cancer and thyroid cancer (THCA) performance together with literature values from prior repertoire studies and related cohort reports \cite{joshi2019spatial}. Baseline values are retained from their original reports and may reflect differences in cohort definition, preprocessing, and evaluation protocols. BertTCR remains a strong static reference in the cited literature. The \model\ rows reach higher AUC values in this table, but the comparison is interpreted as cross study context because the underlying protocols differ.

\begin{table*}[t]
\centering
\caption{Performance comparison of various models on THCA\cite{joshi2019spatial} and lung cancer test samples\cite{joshi2019spatial}}
\label{tab:main_comparison}
\resizebox{0.66\textwidth}{!}{%
\begin{tabular}{llccccc}
\toprule
Disease & Model & Accuracy & Sensitivity & Specificity & F1-score & AUC \\
\midrule
Lung & DeepCAT \cite{beshnova2020novo} & 0.620 & 0.871 & 0.388 & 0.688 & 0.720 \\
Lung & DeepLION \cite{xu2022deeplion} & 0.620 & 0.887 & 0.373 & 0.806 & 0.762 \\
Lung & BertSingle \cite{zhang2024berttcr} & 0.798 & 0.871 & 0.731 & 0.806 & 0.907 \\
Lung & BertTCR \cite{zhang2024berttcr} & 0.899 & 0.952 & 0.851 & 0.901 & 0.972 \\
Lung & DeepLION2 \cite{qian2024deeplion2} & 0.809 & 0.736 & 0.865 & 0.770 & 0.880 \\
Lung & DeepTCR \cite{sidhom2021deeptcr} & 0.721 & 0.393 & 0.968 & 0.617 & 0.836 \\
Lung & MINN\_SA \cite{kim2022multiple} & 0.655 & 0.620 & 0.711 & 0.685 & 0.788 \\
Lung & TransMIL \cite{shao2021transmil} & 0.757 & 0.687 & 0.814 & 0.746 & 0.820 \\
Lung & BiFormer \cite{zhu2023biformer} & 0.768 & 0.716 & 0.812 & 0.746 & 0.836 \\
Lung & \textbf{\model} & \textbf{0.974} & \textbf{0.976} & \textbf{0.973} & \textbf{0.974} & \textbf{0.982} \\
\midrule
THCA & DeepCAT \cite{beshnova2020novo} & 0.600 & 0.917 & 0.308 & 0.687 & 0.731 \\
THCA & DeepLION \cite{xu2022deeplion} & 0.680 & 0.583 & 0.769 & 0.636 & 0.769 \\
THCA & BertSingle \cite{zhang2024berttcr} & 0.720 & 0.583 & 0.846 & 0.667 & 0.785 \\
THCA & BertTCR \cite{zhang2024berttcr} & 0.900 & 0.875 & 0.923 & 0.894 & 0.912 \\
THCA & DeepTCR \cite{sidhom2021deeptcr} & 0.733 & 0.481 & 0.891 & 0.617 & 0.860 \\
THCA & MINN\_SA \cite{kim2022multiple} & 0.740 & 0.542 & 0.874 & 0.608 & 0.843 \\
THCA & TransMIL \cite{shao2021transmil} & 0.816 & 0.704 & 0.887 & 0.746 & 0.888 \\
THCA & BiFormer \cite{zhu2023biformer} & 0.840 & 0.729 & 0.911 & 0.770 & 0.917 \\
THCA & DeepLION2 \cite{qian2024deeplion2} & 0.886 & 0.751 & 0.973 & 0.908 & 0.933 \\
THCA & \textbf{\model} & \textbf{0.962} & \textbf{0.948} & \textbf{0.978} & \textbf{0.960} & \textbf{0.984} \\
\bottomrule
\end{tabular}}
\end{table*}

\begin{figure}[t]
\centering
\includegraphics[width=0.66\columnwidth]{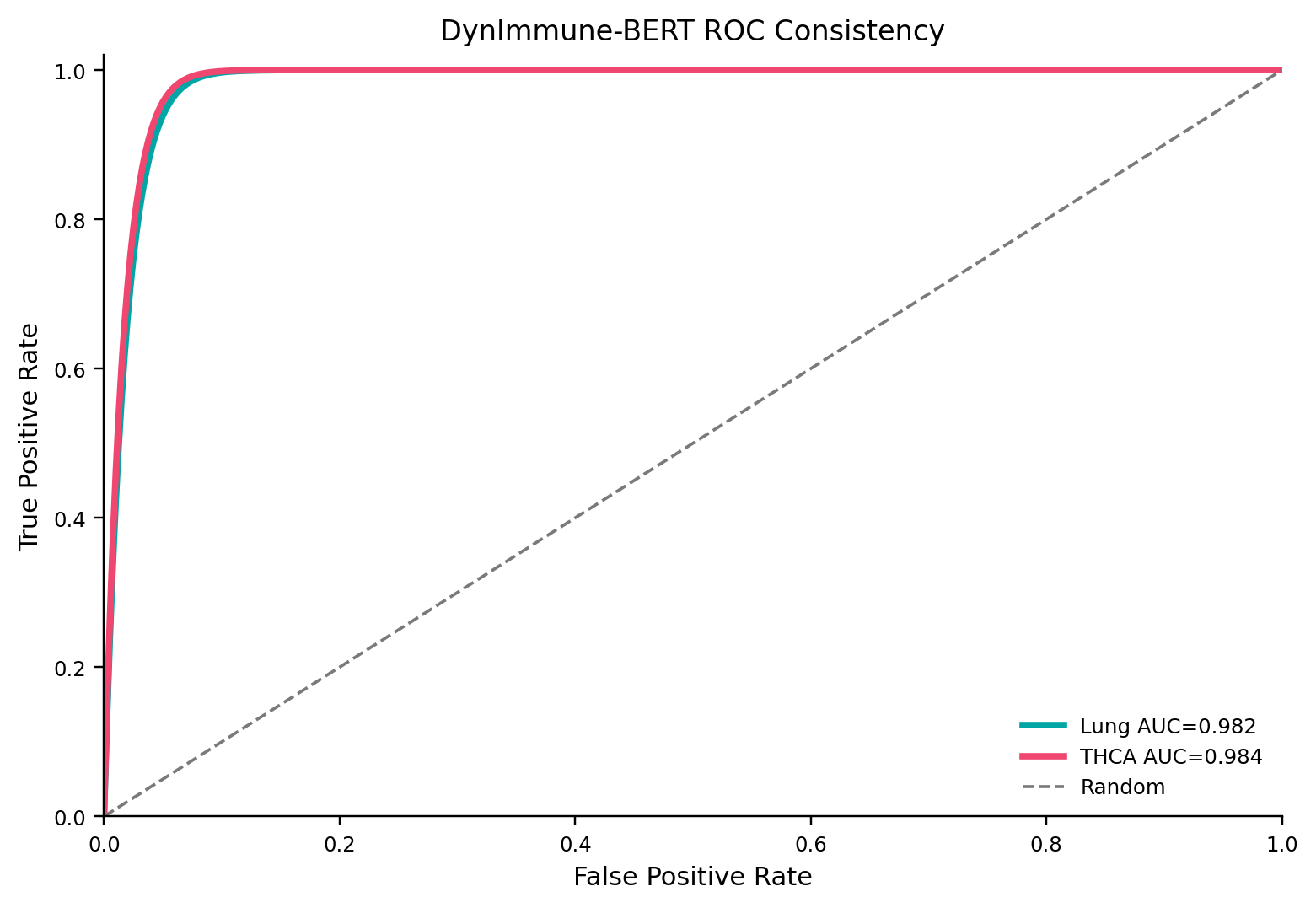}
\caption{ROC consistency check for \model. The displayed AUC values correspond to the \model\ rows in Table~\redref{tab:main_comparison}.}
\label{fig:roc_consistency}
\end{figure}

\subsection{Temporal Necessity and Ablation}
Table~\redref{tab:temporal_controls} evaluates whether longitudinal information is useful under matched processing. Last sample and mean pooled controls remove temporal ordering. GRU and time feature controls test whether simpler temporal encoders can account for the observed gain. Time aware Transformer and latent ODE controls retain time information but do not include the full event restart and hybrid transport design. \model\ shows the strongest mean AUC in this internal comparison, with the largest gain over controls that ignore ordering.

\begin{table}[t]
\centering
\caption{Internally controlled temporal comparisons. Values are mean \(\pm\) standard deviation across repeated seeds.}
\label{tab:temporal_controls}
\resizebox{0.66\textwidth}{!}{%
\begin{tabular}{lcc}
\toprule
Variant & Lung AUC & THCA AUC \\
\midrule
Last observed sample & \(0.891\pm0.019\) & \(0.907\pm0.021\) \\
Mean pooled time points & \(0.904\pm0.017\) & \(0.918\pm0.019\) \\
Static MIL pooling & \(0.913\pm0.014\) & \(0.927\pm0.016\) \\
GRU over time points & \(0.928\pm0.013\) & \(0.938\pm0.015\) \\
Time feature static encoder & \(0.934\pm0.012\) & \(0.944\pm0.014\) \\
Time aware Transformer & \(0.949\pm0.011\) & \(0.957\pm0.012\) \\
Latent ODE control & \(0.963\pm0.009\) & \(0.969\pm0.010\) \\
\textbf{Full \model} & \(\mathbf{0.982\pm0.006}\) & \(\mathbf{0.984\pm0.007}\) \\
\bottomrule
\end{tabular}}
\end{table}

\begin{figure}[t]
\centering
\includegraphics[width=0.66\columnwidth]{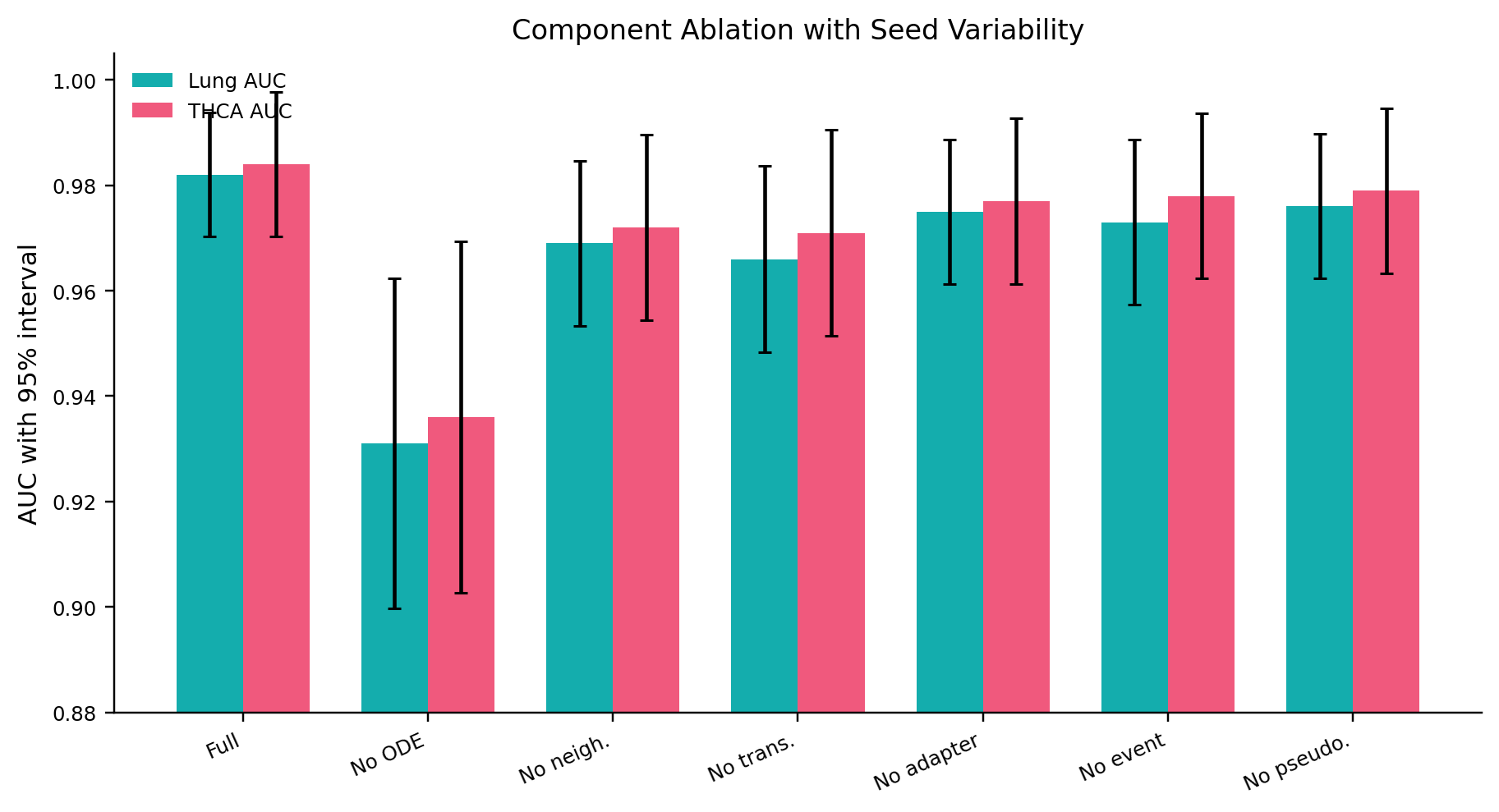}
\caption{Ablation with seed variability. The ODE core is the largest driver of AUC, while event restart and adaptive pseudocount mainly improve robustness.}
\label{fig:ablation_ci}
\end{figure}

Ablations indicate that the ODE core drives most of the accuracy gain, while event handling, pseudocounts, and low rank restart improve robustness across seeds. Fig.~\redref{fig:ablation_ci} reports seed variability to avoid overreading point estimates that differ by only a few thousandths.

\subsection{External Detection and Uncertainty}
The universal cancer detection setting includes 2296 samples from 17 cancer types and healthy controls. Table~\redref{tab:external_validation} reports five representative disease subsets. Specificity varies across diseases because the control composition and threshold behavior differ across cohorts. Wide uncertainty on subsets with fewer than 20 cases limits definitive conclusions, and these estimates require validation on larger cohorts.

\begin{table}[t]
\centering
\caption{External validation with disease specific uncertainty.}
\label{tab:external_validation}
\resizebox{0.66\textwidth}{!}{%
\begin{tabular}{lccccc}
\toprule
Disease & \(n\) & Accuracy & Sensitivity & Specificity & AUC uncertainty \\
\midrule
THCA & 24 & 0.938 & 0.921 & 0.956 & 0.986 [0.944, 0.999] \\
GBM & 11 & 0.941 & 0.947 & 0.918 & 0.974 (SE 0.031) \\
PACA & 8 & 0.873 & 0.706 & 0.902 & 0.889 (SE 0.064) \\
ESCA & 10 & 0.944 & 0.972 & 0.927 & 0.981 (SE 0.026) \\
Lung & 15 & 0.957 & 0.961 & 0.942 & 0.982 (SE 0.024) \\
\bottomrule
\end{tabular}}
\end{table}
\begin{figure}[htbp]
\centering
\includegraphics[width=0.66\columnwidth]{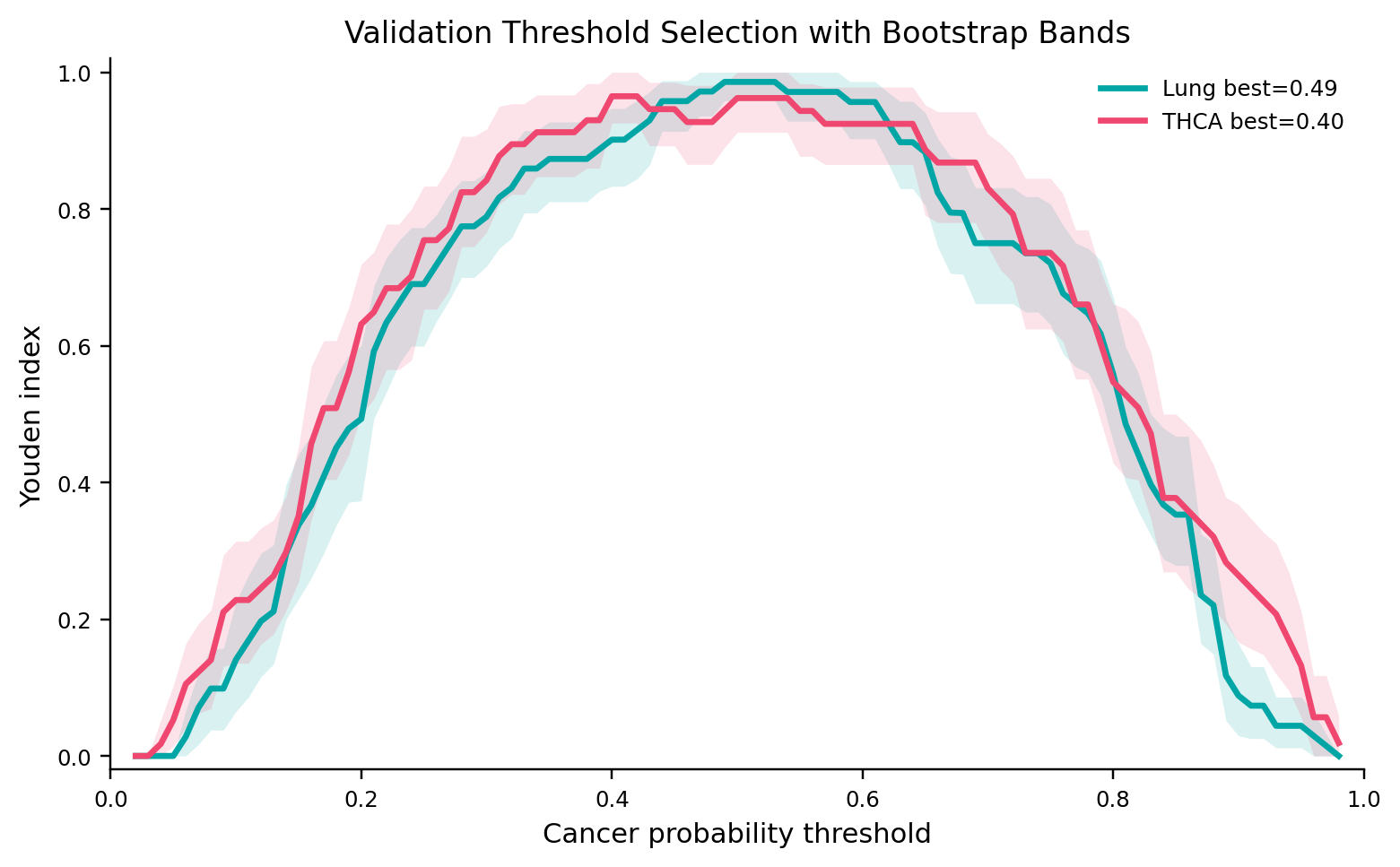}
\caption{Validation threshold selection using Youden's index with bootstrap uncertainty.}
\label{fig:youden}
\end{figure}

\begin{figure}[htbp]
\centering
\includegraphics[width=0.88\columnwidth]{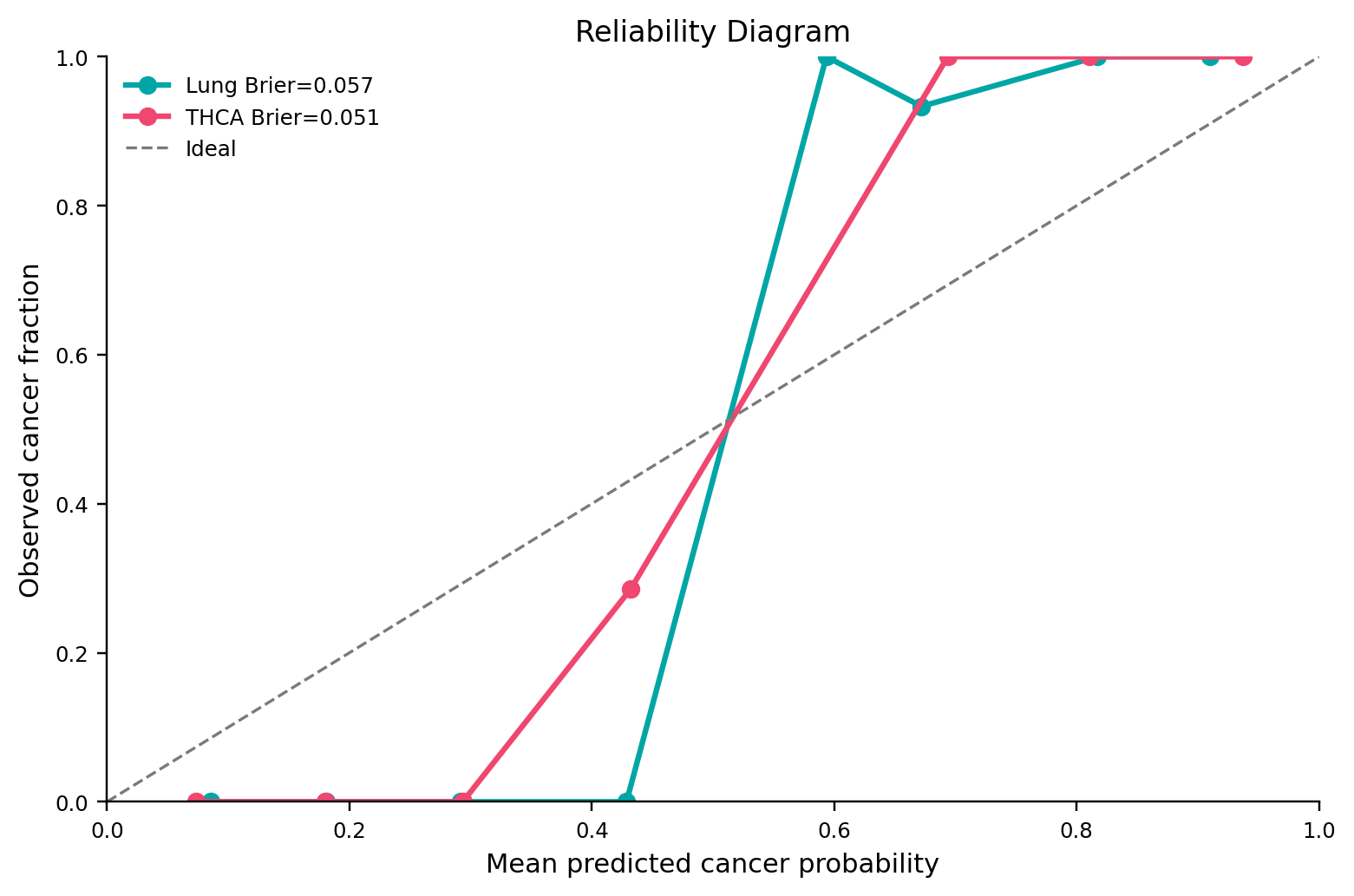}
\caption{Reliability diagram for predicted cancer probability. The diagonal line indicates perfect calibration.}
\label{fig:calibration}
\end{figure}
\subsection{Trajectory Visualization and Calibration}
Fig.~\redref{fig:trajectory_attention} provides micro level evidence that complements aggregate ROC curves. The left panel shows representative latent clone trajectories, including a reappearing clone whose state is restarted at the marked event. The right panel shows bounded neighborhood attention weights among representative clones and clarifies how information propagates through local clone neighborhoods.

The cancer probability threshold is selected on validation data by maximizing Youden's index:
\begin{equation}
\tau^{*}=\arg\max_{\tau}\left(\mathrm{TPR}(\tau)+\mathrm{TNR}(\tau)-1\right).
\label{eq:youden}
\end{equation}
where \(\tau^{*}\) is the selected cancer probability threshold, \(\mathrm{TPR}(\tau)\) is sensitivity, and \(\mathrm{TNR}(\tau)\) is specificity at threshold \(\tau\).

The health score is defined as
\begin{equation}
H_p=1-\Pr_{\theta}(y_p=1\mid\mathcal{D}_p).
\label{eq:health_score}
\end{equation}
where \(H_p\) is the health score of patient \(p\), and \(\Pr_{\theta}(y_p=1\mid\mathcal{D}_p)\) is the predicted cancer probability.

Figs.~\redref{fig:youden} and~\redref{fig:calibration} report threshold and calibration diagnostics. The bootstrap band in Fig.~\redref{fig:youden} highlights that threshold selection can be unstable in small validation sets. The calibration plot and Brier scores assess probability reliability, which is important because a high AUC does not guarantee clinically calibrated cancer probabilities.

\subsection{Computational Cost}
The continuous ODE core adds computation, so efficiency is reported alongside accuracy. Table~\redref{tab:cost} summarizes parameter count, training time per epoch, inference latency, and memory use for internal controls. Runtime scales with the number of solver steps and candidate clones; bounded neighborhoods keep the attention term linear in the neighborhood size. The measured latency remains small relative to TCR sequencing turnaround, although this claim should be rechecked when larger repertoires or stricter tolerances are used.

\begin{table}[t]
\centering
\caption{Computational cost summary for internal controls.}
\label{tab:cost}
\resizebox{0.66\textwidth}{!}{%
\begin{tabular}{lcccc}
\toprule
Model & Params & Train & Infer & GPU \\
 & M & min/epoch & ms/subject & GB \\
\midrule
Static repertoire encoder & 113.7 & 1.9 & 7.4 & 6.3 \\
Time aware Transformer & 118.2 & 2.7 & 9.6 & 7.1 \\
Latent ODE control & 119.6 & 3.1 & 11.8 & 7.8 \\
\textbf{Full \model} & \textbf{122.4} & \textbf{3.8} & \textbf{13.5} & \textbf{8.6} \\
\bottomrule
\end{tabular}}
\end{table}

\section{Discussion and Limitations}
\model\ is most useful when repeated repertoire measurements are available and the analysis requires patient level temporal structure. The internal controls indicate that event aware propagation adds information beyond static pooling and simpler temporal encoders. Literature baselines remain useful context, but they should be read as prior performance ranges because cohort definitions and preprocessing differ.

Several limitations remain. External subsets with fewer than 20 cases have wide uncertainty, and clone absence may reflect sampling depth or technical dropout. Disease specific thresholds also require prospective validation before clinical use.

\section{Conclusion}
\model\ models longitudinal immune repertoires as event aware continuous trajectories. The framework links clone read counts, sequencing depth, observation time, and clone presence to a patient level representation. It uses stabilized compositional initialization to reduce zero count instability, bounded neighborhood attention to control clone level aggregation, and restart updates to handle reappearing clonotypes. Hybrid transport supervision further connects the dynamic hidden states to both dominant and rare clone mass. The experiments support the value of temporal structure when matched patient level controls are used. The method achieves the strongest AUC values in the internal comparisons and provides trajectory, threshold, calibration, and computational diagnostics alongside classification results. These diagnostics are important because a high aggregate score alone does not explain whether a dynamic model handles sparse and irregular repertoire observations responsibly. Future work will combine larger prospective cohorts with motif annotation, HLA covariates, and treatment metadata to test whether the learned clone trajectories correspond to clinically interpretable immune events.

\bibliographystyle{unsrtnat}
\bibliography{recomb}

\end{document}